\documentclass[runningheads]{llncs}
\usepackage{amsmath}%
\usepackage[T1]{fontenc}
\usepackage{graphicx}
\usepackage{booktabs}
\usepackage[misc]{ifsym}
\usepackage{hyperref}

\usepackage{comment}
\usepackage{graphicx}%
\usepackage{multirow}%
\usepackage[title]{appendix}%
\usepackage{xcolor}%
\usepackage{textcomp}%
\usepackage{manyfoot}%
\usepackage{booktabs}%
\usepackage{algorithm}%
\usepackage{algorithmicx}%
\usepackage{algpseudocode}%
\usepackage{listings}%
\usepackage{subcaption}

\usepackage{mwe}

\begin{document}





\title{Scaling Time Series Classification via XAI-Driven Data Reduction}

\titlerunning{XAI-Driven Data Reduction}

\author{Davide Italo Serramazza \Letter \and Thach Le Nguyen \and Georgiana Ifrim}

\tocauthor{Davide Italo Serramazza, Thach Le Nguyen, Georgiana Ifrim}

\authorrunning{Serramazza et al.} 
%
\institute{School of Computer Science, University College Dublin, Ireland \\
\email{davide.serramazza@ucdconnect.ie}\\
\email{\{thach.lenguyen,georgiana.ifrim\}@ucd.ie}}

\maketitle              

\begin{abstract}
Explainable AI (XAI) for time series has seen significant algorithmic growth, but its utility in providing measurable performance gains for downstream tasks remains under-explored. This paper bridges this gap by introducing drXAI, a novel methodology that repurposes XAI attribution methods for effective data reduction in Time Series Classification (TSC).
The core challenge in modern TSC is scalability; state-of-the-art models, such as Transformers, exhibit quadratic complexity relative to sequence length and linear complexity relative to the number of channels. This renders them computationally prohibitive for massive datasets. drXAI addresses this by using a fast, GPU-accelerated classifier (Hydra) to generate local attributions. We aggregate these into global feature importance scores and employ an automated elbow-cut heuristic to select the most salient features without requiring manual thresholds.
We evaluate our approach on both synthetic and real-world univariate and multivariate datasets. On synthetic benchmarks, drXAI successfully recovers ground-truth features where traditional baselines fail. On real-world data, drXAI achieves 80–90\% data reduction while maintaining classification accuracy comparable to models trained on the full dataset. Most importantly, we show that drXAI allows resource-intensive models like ConvTran to scale to datasets that were previously inaccessible due to memory constraints. Our results show the benefits of using XAI not just for interpretability, but as a robust tool for feature selection and scalability in time series analysis. All our code and data are openly available.
\end{abstract}


\section{Introduction}

The field of Explainable AI (XAI) has experienced significant growth in recent years, particularly within the Time Series Classification (TSC) domain. A major focus of XAI is feature attribution, which quantifies the importance of input features for the model predictions. This area has seen considerable advances in the efficiency and effectiveness of attribution methods for time series data \cite{theissler2022}. 

Despite these algorithmic advancements, the utility of XAI as a tool for achieving measurable performance gains in other computational tasks remains underexplored. Our work offers a paradigm shift: viewing XAI not only as interpreting complex models but also as a practical tool for enhancing model efficiency and scalability. This paper bridges this gap by introducing \textbf{drXAI (Data Reduction with XAI)}, a novel attribution-agnostic methodology that repurposes attribution methods for effective feature selection in TSC. 
For many State-of-the-Art (SOTA) TSC models, the time complexity \textbf{grows linearly with the number of features or even quadratically} with sequence length (e.g., transformers); additionally, these models also require a huge amount of memory. This computational burden is often a limitation for training on massive, high-dimensional datasets. A way to scale these models is to train them on a reduced, yet highly informative, set of features. Our work directly enables this by using XAI to identify the most important features, thus reducing data dimensionality without sacrificing critical information. To our knowledge, this work is the first to use XAI for feature selection to scale time series classification methods.

drXAI is a \textit{wrapper} feature selection method that operates in two stages: it first trains a TSC model, referred to as the \textit{explainer classifier}, and then generates explanations for its predictions. In this work, accounting for speed, we use the GPU implementation of Hydra \cite{dempster2024highly} as the explainer classifier, restrict attribution to the \textbf{explainer set}, a subset of the training set, and apply lightweight explainers such as Feature Ablation.

For Multivariate Time Series Classification (MTSC) datasets, we focus on \emph{channel selection}. We propose a fast approach to compute global channel importance by aggregating the local attribution values, ranking the channels, and selecting a subset. We then retrain SOTA classifiers on the reduced data and measure the accuracy and computational gains. For UTSC datasets, we apply the same algorithm for computing global time point importance, which is then used for data reduction via consecutive \emph{time point selection}.
This is especially effective for very long time series, where SOTA TSC methods do not scale well due to the requirement of extensive computational resources. 

\textbf{Our main contributions in this paper are:}
\begin{enumerate}
    \item We develop \textbf{drXAI}, a general framework that leverages XAI attribution to select features from MTSC and UTSC datasets. The algorithm is attribution-agnostic and is general enough to support both channel and time-point selection.
     As part of this, we also investigate the \textit{background data} used for simulating data missingness in attribution methods and propose \textit{Proto}, a single-instance background outperforming the standard zeros baseline common in popular attribution methods  \cite{kokhlikyan2020captum}.
   
    \item   Using synthetic MTSC data, we show that drXAI selects only informative channels, unlike baseline methods. On synthetic UTSC data, it successfully selects over 90\% of relevant features, whereas baselines are limited to at most 33\%. On real-world datasets, drXAI provides the best trade-off between data reduction and accuracy, most of the time matching the models trained on all features, for both MTSC and UTSC experiments.

    \item We demonstrate that by using a fast classifier (Hydra) and a fast attribution method (Feature Ablation) for data selection, we can effectively train more accurate but resource-intensive models (e.g., ConvTran, MultiRocket-Hydra  \cite{middlehurst2023bake}). This enables computationally expensive models to train successfully, overcoming the bottleneck in use cases where the model cannot be trained on the entire dataset due to memory limits. 
    
    \item To encourage further research and reproducibility in this area, we make all our data and code publicly available\footnote{\url{https://github.com/mlgig/drXAI}}.
\end{enumerate}

\section{Background}
\label{sec:background}

\subsection{Problem Definition}
We represent a time series dataset $D$ as a tensor, 
of dimensions $n \times d \times L$ where $n$, $d$, and $L$ respectively represent the number of samples, channels, and time points in the data.  In the UTS case, $d=1$.

A (trained) classification model $clf$ predicts the class $c_i$ of a time series instance $D_i$ ($1 \le i \le n$). An explainer $exp$ explains the $clf$ predictions on $D$ by producing a set of attribution maps $\mathcal{A}$, a tensor of attributions with the same dimensions as $D$ ($n \times d \times L$) where $\mathcal{A}_{i,j,k}$ indicates the attribution (i.e., relevance) of $D_{i,j,k}$ for the prediction of the model $clf$ on the instance $D_i$.

We denote the set of features as ${F}=\{f_1, f_2, \dots, f_m\}$ where $m$ is the total number of features. Each feature represents a portion of the time-series data, e.g., data within the same time segment, the same time step, or the same channel. The union of all features is the complete time series. We focus on the task of selecting a subset of features $F_{sel} \subset F$. 
We informally refer to this new set as \textit{selected features}, denoting as $D_{F_{sel}}$ the new dataset containing only these features.

In this work, we focus on data reduction for TSC, organised into two subtasks: the \emph{channel selection} problem for MTSC datasets and \emph{time points selection} for UTSC datasets. For the former, each feature represents a channel ($m=d$) while for the latter, each feature represents a time point ($m=L$). To avoid confusion, we specify which type of features we focus on when describing specific data reduction tasks.


\subsection{Time Series Classification}
\label{subsec:classifiers}

Recent work in TSC has significantly advanced the availability of extensive TSC benchmarks \cite{uea-mtsc-archive,dempster2025monstermonashscalabletime} as well as the accuracy and efficiency of TSC algorithms \cite{foumani2024improving,dempster2024highly}.
While many algorithms achieve top accuracy on benchmarks, they are resource-intensive, especially for large-scale datasets or very long time series \cite{middlehurst2023bake}. We discuss a few relevant SOTA classifiers, as well as their computational complexity.



\textbf{Explainer Classifier.} Our proposed method, drXAI, is a wrapper method for feature selection and requires a fast classifier to be explained. 
\textbf{Hydra} is a TS \textit{transformation} algorithm combining convolution-based and dictionary-based aspects: $g$ groups of $k$ convolutional kernels slide over the TS, and for each group, the closest-matching kernels are counted at each time-point. Features are then fed to a Ridge Classifier.  
The time complexity of Hydra is dominated by the convolution and the competitive counting process.
Since k is usually a fixed constant, the time complexity is effectively linear $O(ndL)$. Hydra's memory complexity is 
$O(nk+kd)$.
In our experiments, we used the fast Hydra GPU implementation from \cite{dempster2024highly}, which scales well to very large datasets. 

\textbf{SOTA Time Series Classifiers.} To assess the quality of data selection $F_{sel}$, we trained the following 3 SOTA classifiers on the reduced datasets $D_{F_{sel}}$.

\textbf{MultiRocket-Hydra (MRH)} \cite{middlehurst2023bake} uses the Hydra features concatenated with those from MultiRocket \cite{tan2022multirocket}. MultiRocket applies kernels also to the first-order difference of the series and extracts features via 4 pooling operators (MPV, MIPV, LSPV, PPV). \\
The MultiRocket pipeline is heavier computationally than Hydra. While still linear in the inputs, the constants for MultiRocket are larger, as well as the number of features extracted (by default around 50,000 features), which makes it more expensive both time and memory wise, especially for long time series.
 Moreover, since by default each kernel is applied to at most 8 random channels, both algorithms also benefit from selections shorter than 8 channels. Beyond efficiency, restricting the input to informative features can further improve model robustness and accuracy.

\textbf{ConvTran} \cite{foumani2024improving} is a recent transformer model, tailored for MTSC. It extracts features from the raw series using convolutional layers that are then fed into the transformer after tokenization. 
ConvTran comes with a significant jump in complexity as compared to Hydra. While Hydra uses random kernels and a linear classifier, ConvTran is a fully trainable Transformer-CNN hybrid that utilizes self-attention.
ConvTran’s complexity is driven by the quadratic nature of the attention mechanism and the parameters of its convolutional embedding layers. Its cost is quadratic w.r.t the series length $O(nL^2d_{embedding})$ and linear w.r.t the number of channels.  
For very long sequences where $L>1000$, this becomes significantly slower than Hydra's $O(nL)$.

\textbf{InceptionTime} \cite{ismail2020inceptiontime} is another deep learning classifier, specifically, an ensemble method composed of 5 vanilla Inception networks (CNN). Its complexity is primarily determined by its deep \textit{Inception} modules, which apply parallel convolutions of varying kernel sizes through bottleneck layers, capturing both local and long-range temporal relationships. 
Unlike ConvTran, InceptionTime is linear with respect to sequence length and channels, but since it requires backpropagation, the constant factors and hardware requirements are much higher than those of Hydra.

\subsection{XAI Methods for Time Series Classification}
\label{subsec:explainers}

Our methodology is \textit{attribution-method agnostic}, i.e., it requires only an attribution map as input, regardless of which algorithm computed it. In this work, we select two \textbf{permutation-based} attribution methods that were adapted for TS and work efficiently in this domain \cite{serramazza2024improving,turbe2023evaluation}. These methods require a \textit{background set} to simulate data missingness when computing the attribution. 
No \textbf{gradient-based} attribution methods were considered for this study primarily because they can't work with non-gradient-based algorithms, like Hydra and MultiRocket.
For the rest of this paper, we use \textbf{explainers} to refer to the XAI attribution methods.

\textbf{Shapley Value Sampling (SVS)} is an  approximation of the \textit{Shapley values} as described in the original work \cite{lundberg2017unified}. It applies sampling to the SHAP formula, randomly permuting the features to be explained, adding them sequentially to each sample in the background set. The attribution values are the change in the model output resulting from these substitutions.  
Although shown to be effective with regard to pointing out important features \cite{serramazza2023evaluating}, this method requires a vast amount of computation time, due to the high number of feature permutations. Some adaptations for time series, such as grouping features through TS segmentation, have been shown to drastically reduce computation time, while preserving accuracy \cite{serramazza2024improving}.

\textbf{Feature Ablation (FA) \cite{kokhlikyan2020captum}} is much simpler and faster when compared to SHAP. It sequentially replaces each feature with the corresponding values of the samples found in the background dataset. As for SVS, the attribution values are the differences in the model output after substituting feature values.

We used the implementations provided in \cite{serramazza2024short} for both explainers.

\subsection{Feature Selection for TSC}
\label{subsec:data-reduction}

We first discuss recent work on \textbf{channel selection for MTSC}. In \cite{dhariyal2023scalable}, the authors propose a supervised algorithm for selecting a subset of channels for MTSC datasets. The algorithm has two variants, ECS and ECP, where the channels are selected based on their discriminative power, estimated using the Euclidean distance between class centroids: a higher distance implies higher discriminative power. 
According to the original paper, this \textbf{filter} method can reduce, on average, 70\% of the data without compromising the accuracy of classifiers. \\
Another recent work that directly compares to ECS and ECP is TSelect \cite{nuyts2025tselect}, a \textbf{wrapper} method which trains for each channel a Logistic Regression classifier based on 5 computationally cheap features extracted channel-wise. These models are then used to determine which of the relative channels to retain: various filters based on these model accuracies, redundant predictions, etc., select which channels to discard.

Reduction of time series data can also be performed by projecting the high-dimensional time series to a lower-dimensional latent space \cite{data-reduction-survey}. Known methods include Autoencoders, Principal Component Analysis, Singular Value Decomposition, Discrete Fourier/Wavelet Transform, and Down-sampling. However, these are \textbf{dimensionality reduction techniques} that project time series into a different space, losing the original features.  
For many applications, it is important to keep the data in the original representation to be able to audit important features, for example, medical applications such as monitoring human health. 
Therefore, these transformation methods are outside the scope of this paper. 

For \textbf{time point selection techniques for UTSC}, a simple and effective approach, especially for long time series, is to use \textit{Random Forest importance} (RFI), using the importance (impurity reduction) of each feature computed during training. Another alternative is the \textit{Mutual Information} (MI) between each feature and targets for a TSC task \cite{scikit-learn}.

\section{Proposed Methodology}
\label{sec:method}

In this section, we describe \textbf{drXAI}, our algorithm for time series data reduction using XAI attribution methods to identify and select important features from TSC datasets.

\begin{algorithm} [h]
	\caption{drXAI Algorithm to Select a Feature Subset from a Time Series Dataset} 
	\begin{algorithmic}[1]
    	\Require $clf$, $exp$, $D$, $n_{samp}$  \space  \Comment{explainer classifier, explainer, train set, n. samples per class}

        \State Train $clf$ on $D$

        \State $D_{exp}$ = sample $n_{samp}$ per class
        
        \State $\mathcal{A}=$ \space Explain $clf$ classifications for samples in $D_{exp}$ using $exp$ 

        
                                
        \State ${A}=$ equation (\ref{eq:channel_sel}) if channel selection else equation (\ref{eq:timePoints_sel})   \Comment{ details in Section \ref{sec:avg_along_dim}  }       
        
        \State    $fr_{avg}=$ \space aggr\_avg(${A})$  \space ; \space
        $fr_{abs}=$ \space aggr\_abs(${A})$ \Comment{ details in Section \ref{sec:aggragate_over_samples} }

        
        \State $F_{sel\_avg}=elbow\_cut(fr_{avg})$ \space ; \space $F_{sel\_abs}=elbow\_cut(fr_{abs})$  
        \Comment{ details in Section \ref{sec:elbow_cut_howUse}  }

        \State  $ F_{sel} = F_{sel\_abs} \cap F_{sel\_avg} $
        \State return $F_{sel}$
	\end{algorithmic} 
    \label{alg:pseudocode}
\end{algorithm}

Our methodology (Algorithm \ref{alg:pseudocode}) relies on the explanation set $D_{exp}$, which is a subset of the training set (assuring there is no information leakage), composed by randomly sampling $n_{samp}$ per class, the explanation classifier $clf$ i.e., the model which is explained (in this work Hydra) using the explainer $exp$. The key functions are:
\begin{itemize}
    \item The \textit{aggr\_avg} and \textit{aggr\_abs} functions, which aggregate the attribution values of different samples based on two complementary strategies.
    \item \textit{Elbow cut} which selects the top $K$ features $F_{sel}$. 
\end{itemize}

\subsection{Computing Feature Attributions}
\label{sec:avg_along_dim}

\begin{figure*} [t]
    \centering
    \includegraphics[width=1.0\linewidth]{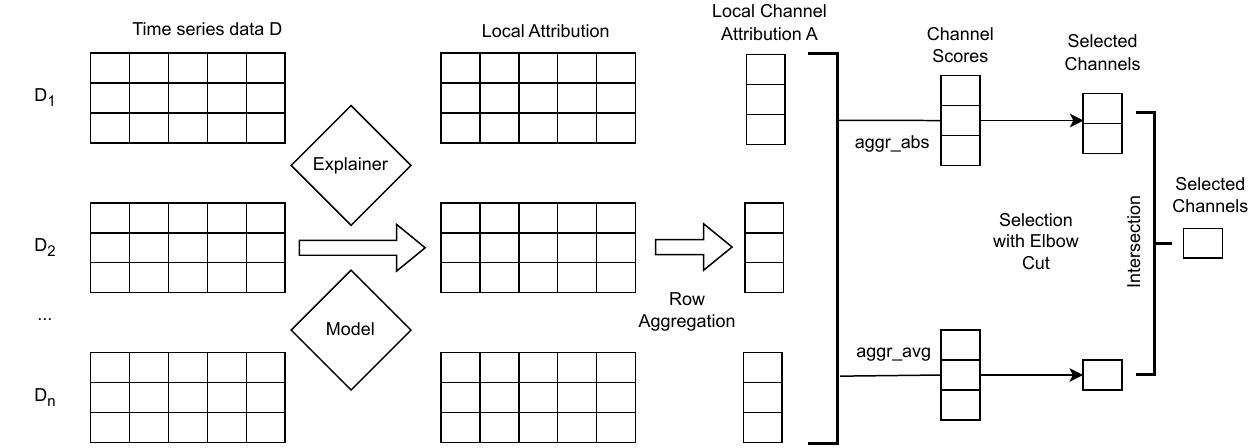}
    \caption{drXAI: Channel selection for MTSC using XAI scores computed from time series attributions.}
    \label{fig:chan_sel_xai}
\end{figure*}

After training $clf$ using the training set $D$ and instantiating the explanation set $D_{exp}$, the result of applying the explainer $exp$ to the classifier $clf$ and the explanation set $D_{exp}$ are the \textit{local attributions} $\mathcal{A} \in \mathcal{R}^{n \times d \times L} $ where $n=|D_{exp}|$. The next step is to calculate the \textbf{feature attribution} of the feature set $F$ for each time series in $D_{exp}$. The attribution of a feature is simply the average attribution of all data points represented by the feature. In particular, if features represent channels (for channel selection), the attribution of a feature is:

 \begin{equation}
    {A}_{i,j} = \frac{1}{L} \sum_{t=1}^L \mathcal{A}_{i,j,t}  \label{eq:channel_sel}
 \end{equation}

where ${A}_{i,j}$ is the attribution of channel $j$ (or feature $F_j$) in time series $D_i$. This is the row aggregation step in Figure \ref{fig:chan_sel_xai}.

Similarly, if each feature represents a time point (for time point selection), the feature attribution is:

\begin{equation}
  {A}_{i,t} = \frac{1}{d} \sum_{j=1}^d \mathcal{A}_{i,j,t}    \label{eq:timePoints_sel}
\end{equation}

where ${A}_{i,t}$ is the attribution of time point $t$ (or feature $F_t$) in time series $D_i$.
Stacking the attributions for each feature and each sample, we obtain the matrix $A$.
To be noted that the next steps are executed regardless of the selection type (channel or time points).

\subsubsection*{Attribution Background Set.} \label{subsec:backgrounds} As mentioned in Section \ref{subsec:explainers}, both  explainers require a \textit{background dataset}. Since the computational complexity of explaining linearly increases with the cardinality of this set, we study two different single-sample backgrounds $b$. \\

\noindent\textbf{Zeros background (Zeros).} A default choice for most explanation frameworks, i.e., a time series full of zeros. Although this is conceptually very simple, it is an unrealistic sample, potentially leading to unreasonable explanations. Mathematically, this is defined as: 
\begin{equation*}
    b =\mathbf{0}_{d \times L}
\end{equation*}


\noindent\textbf{Class prototypes average (Proto).} The background we propose uses the prototype of each class $c$:  
\begin{equation}
       p_c = \frac{1}{| D^c |} \sum_{D_i \in D^c} D_i
\end{equation}

where $D^c \subset D$ is the set of all samples in the training set, belonging to class $c$. Let $C$ be the set of all classes. The proto background is defined as the average of class prototypes:
\begin{equation}
    b = \frac{1}{|C|} \sum_{c \in C} p_c
\end{equation}

\subsection{Aggregating Feature Attribution over Samples}
\label{sec:aggragate_over_samples}

The previous step computes the sample-wise attributions for each feature in the feature set F. The next step is to aggregate these local attributions over the explanation set $D_{exp}$ to obtain  \textbf{global attributions}. This results in the feature importance scores $\mathbf{fr_{avg}}, \mathbf{fr_{abs}} \in \mathcal{R}^{m}_{+}$, which can be used for feature selection (e.g., this is the channel scores vector in Figure \ref{fig:chan_sel_xai}). 
We use two complementary approaches for this aggregation which proved effective in our experiments.

\textbf{aggr\_avg}: In this scenario, the feature importance score is averaged over all samples first, then the absolute value is computed. This strategy aims to de-emphasize uncertain features, i.e., features that have mixed negative and positive attribution signs (thus roles) across the dataset.
\begin{equation}
fr_{avg} = | \frac{1}{n} \sum_{i=1}^n A_{i}|
\end{equation}

where $A_i$ is the $i$-th a row of $A$, i.e., attribution vector of the sample $D_i$. \\
\textbf{aggr\_abs}: On the other hand, this strategy averages the absolute value of the attributions for each feature: 
\begin{equation}
    fr_{abs}=  \frac{1}{n} \sum_{i=1}^n | A_{i}|
\end{equation}

This strategy simply detects the most \textit{active} features, regardless of the sign. The dimension of these vectors is  $m=d$ for channel selection and $m=L$ for time point selection.


\subsection{Feature Selection using Elbow Cut}
\label{sec:elbow_cut_howUse}
The \textit{elbow cut} of a curve is a common heuristic to choose a point where intuitively the diminishing returns are no longer worth the additional cost; e.g., it is often used to select the number of clusters while running \textit{k-means} algorithms, and in the channel selection context for the ECP and ECS methods as described in \cite{dhariyal2023scalable}.

In drXAI, the \textit{elbow cut} is used after all features are sorted by their computed score, to automatically select the number of top features to retain: $F_{sel\_abs}$ and $F_{sel\_avg}$ are respectively the results of the elbow cut applications on $fr_{abs}$ and $fr_{avg}$. 
The set of final selected features $F_{sel}$ is the intersection of $fr_{abs}$ and $fr_{avg}$ (Line 7 Algorithm \ref{alg:pseudocode}). Intuitively, this set contains only features that are both magnitude- and sign-wise important, thereby including only the essential ones.  

\subsection{Time and Space Complexity for drXAI}
\label{subsec:complexity}

The computational efficiency of the \textbf{drXAI} framework is a primary contribution, specifically designed to mitigate the prohibitive costs of training classifiers on high-dimensional time series. The total complexity of the pipeline is the sum of three distinct phases: (i) Explainer Classifier Training, (ii) Attribution Generation, and (iii) Feature Selection.

\noindent\textbf{Explainer Classifier Training.} 
We use \textbf{Hydra} as the core classifier for its linear-scaling properties. For a dataset $D(n, d, L)$, Hydra extracts features using $k$ random convolutional kernels 
Since $k$ is a fixed hyperparameter, this phase remains $O(n \cdot d \cdot L)$, which is asymptotically optimal for time series processing.

\noindent\textbf{Attribution Generation.} 
The complexity of this phase is the product of the number of samples to be explained and the cost of the chosen explainer:
\begin{itemize}
    \item \textbf{Feature Ablation (FA):} For each sample, FA requires a forward pass for each feature. For channel selection, this is $O(d \cdot \text{Cost}_{\text{Hydra}})$, and for time-point selection, $O(L \cdot \text{Cost}_{\text{Hydra}})$.
    \item \textbf{SHAP:} SHAP estimates values via  sampling. While the number of samples $s$ scales with the feature space, Hydra's GPU-accelerated inference, coupled with TS segmentation, allows SHAP to remain tractable even for $L > 1,000$.
\end{itemize}

\noindent\textbf{Feature Selection.} This is a step that has a constant time involving sorting the features by importance and applying the elbow cut.

\textbf{Space Complexity.} The memory footprint is dominated by the storage of attribution maps $O(D_{exp} \cdot d \cdot L)$. However, the subsequent \textit{elbow-cut} selection enables a significant reduction in the memory required for training deep models. As demonstrated in our results (Section \ref{sec:exp}), this reduction is the critical factor enabling \textbf{ConvTran} to run on datasets where it would otherwise trigger Out-of-Memory (OOM) errors.
\section{Experiments}
\label{sec:exp}

In our experiments, we consider 4 configurations of our methods: the combinations of previously listed explainers and backgrounds, SHAP Proto, FA Proto, SHAP zeros, and FA zeros. We hypothesize that the informative Proto background gives a small boost to our method compared to the uninformative zeros. We also hypothesize that, because SHAP is a more complex algorithm, it yields better selection than FA. \\
To assess the quality of the selected features, the following pipeline is applied to each SOTA classifier described in Section \ref{subsec:classifiers}, and each dataset described in Sections \ref{sec:MTSC_exp} and \ref{sec:utsc_exp}.

\begin{itemize}
    \item  Hydra is trained on the current dataset. This allows the application of the drXAI algorithm in the 4 configurations previously listed (e.g., drXAI-SHAP-Proto). Specifically, for each of those, we get the selected features $F_{sel}$.
    \item Each baseline, along with a random selection method, is evaluated on the dataset to obtain its $F_{sel}$.
    \item For each of the $F_{sel}$, we trained each SOTA classifier 3 times using the reduced dataset $D_{F_{sel}}$, recording the mean accuracy and the mean time for training plus inference. Multiple training rounds (3) were done to account for the stability of the trained classifiers. 
    \item Our evaluation of each selection is based on mean accuracy over the 3 runs, and on the percentage of saved data. 
\end{itemize}

Experiments were conducted using a machine with AMD EPYC 9654P CPU (96 cores, 192 threads), NVIDIA GeForce RTX 4090 GPU (24GB VRAM), and 1.5Tb of RAM. 
Considering the breadth of experiments, the main article reports only the most salient results, while the Appendix provides more details.

\subsection{Classifier Training}

For MRH, we implemented our version based on MultiRocket and Hydra transformations from the \emph{aeon} library \cite{aeon24jmlr} and the  RidgeCV Classifier in \emph{sklearn}  \cite{scikit-learn}. For Hydra, we used the default hyperparameters, setting the batch size to 256. For ConvTran and InceptionTime training, we used the strategy of \cite{dempster2025monstermonashscalabletime}, i.e., reserving 10\% of the training set as a validation set. We allow up to 100 epochs with early stopping, using the validation loss as criterion. We set the batch size to 256. 

Finally, for computationally demanding UTSC datasets, we make two adjustments: we reduce the batch size for InceptionTime and ConvTran to fit the GPU memory and switch to an iterative solver for the Ridge Classifier in MultiRocket-Hydra, which is required when its input matrix exceeds a maximum size.

\subsection{Channel Selection for MTSC Datasets}
\label{sec:MTSC_exp}

The MTSC datasets have fewer samples than the UTSC ones (Appendix). Thus, we set $n_{samp}$, the number of sampled instances per class composing $D_{exp}$, to 50. We also set the maximum non-improving epochs before early stopping for ConvTran and InceptionTime to 20.  
drXAI is compared against three recent baselines, ECP, ECS, and TSelect, as well as a random selection baseline that samples both the number of channels and the exact  channels to retain from a uniform distribution.

\subsubsection{Synthetic Dataset.}

\begin{table} [t]
    \centering
    \resizebox{0.99\textwidth}{!}{
    \begin{tabular}{|l|c|c|c|c|c|c|c|c|}
     \hline
     
      & drXAI-FA-Proto & drXAI-SHAP-Proto & drXAI-FA-zeros & drXAI-SHAP-zeros  & ECP & ECS  & TSelect & random \\
     \hline
         n. informative 20  & \textbf{19}  & 9 &  4  & 2 & 12 & 4 &  4 & 2 \\
    \hline
        n. uninformative 20  & \textbf{0} & \textbf{0} & \textbf{0} & \textbf{0} & 17 & 2 &  6 & 5 \\
    \hline
      mean accuracy & \textbf{.735} & .658 & .586 & .555 & .634 & .578 & .555 & 0.536 \\
    \hline
    \end{tabular}}
    \caption{Number of informative/uninformative channels selected for synthetic MTSC data and  mean accuracy (across 3 runs of SOTA classifiers).}
    \label{tab:synth_multi}
\end{table}

For the MTSC synthetic dataset, we used the data generator code from \cite{serramazza2024improving}. Each channel is composed of a lower frequency \textit{sine wave}. For each sample, two higher-frequency \textit{support waves}, shorter than the previous ones and with frequencies varying within a specific range, are injected into two randomly selected informative channels. The binary classification task is whether the sum of these two frequencies exceeds a threshold. \\
We generate 5,000 samples for each of the train and validation sets. The series length is set to 1,000 time points; there are 20 informative and 20 uninformative channels (40 total). Using this controlled dataset, we can evaluate how many of the first 20 informative and the last 20 uninformative channels are selected: the ideal selection is all of the informative, none of the uninformative channels. \\
Table \ref{tab:synth_multi} shows that only our method \textbf{exclusively selects informative channels}; using the Proto baseline, FA selects 19 out of 20 channels, and SHAP selects 9. This results in these two selections achieving the top two accuracies. Specifically, the drXAI-FA-Proto configuration outperforms the best baseline (ECP) by 10 percentage points in mean accuracy.

\subsubsection{Real-world Data.}

\begin{figure} [t]
    \centering
    \includegraphics[width=1.1\linewidth]{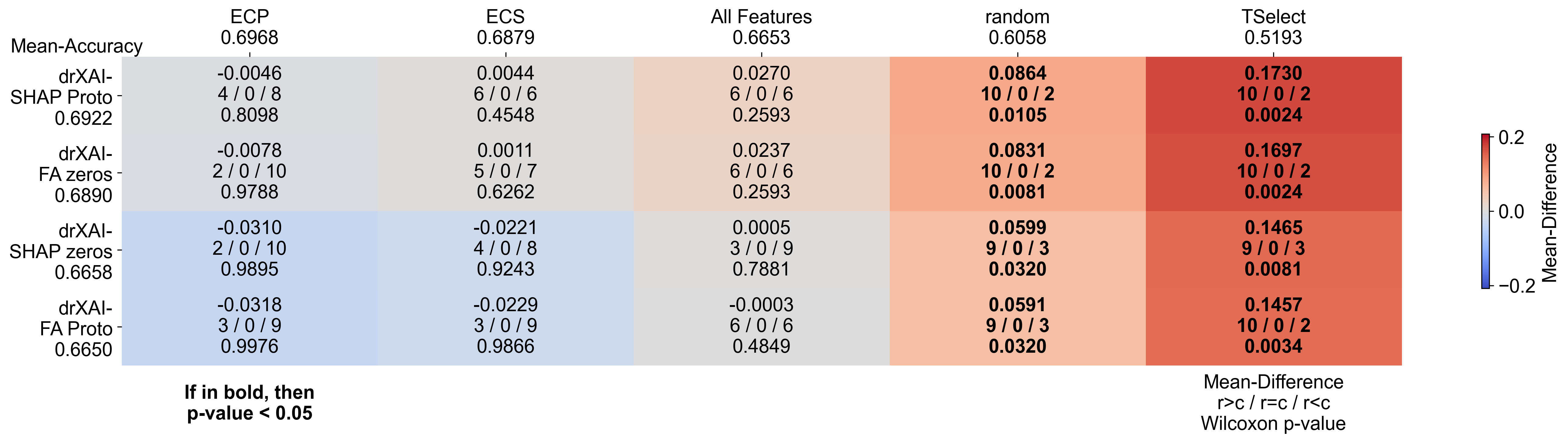}
    \caption{Mean accuracy of each selection (and All Features) for the 4 MTSC real-world datasets and the 3 SOTA classifiers, yielding 12 results in total.}
    \label{fig:multi_acc}
\end{figure}

\begin{figure} [htb]
    \centering
    \includegraphics[width=1.1\linewidth]{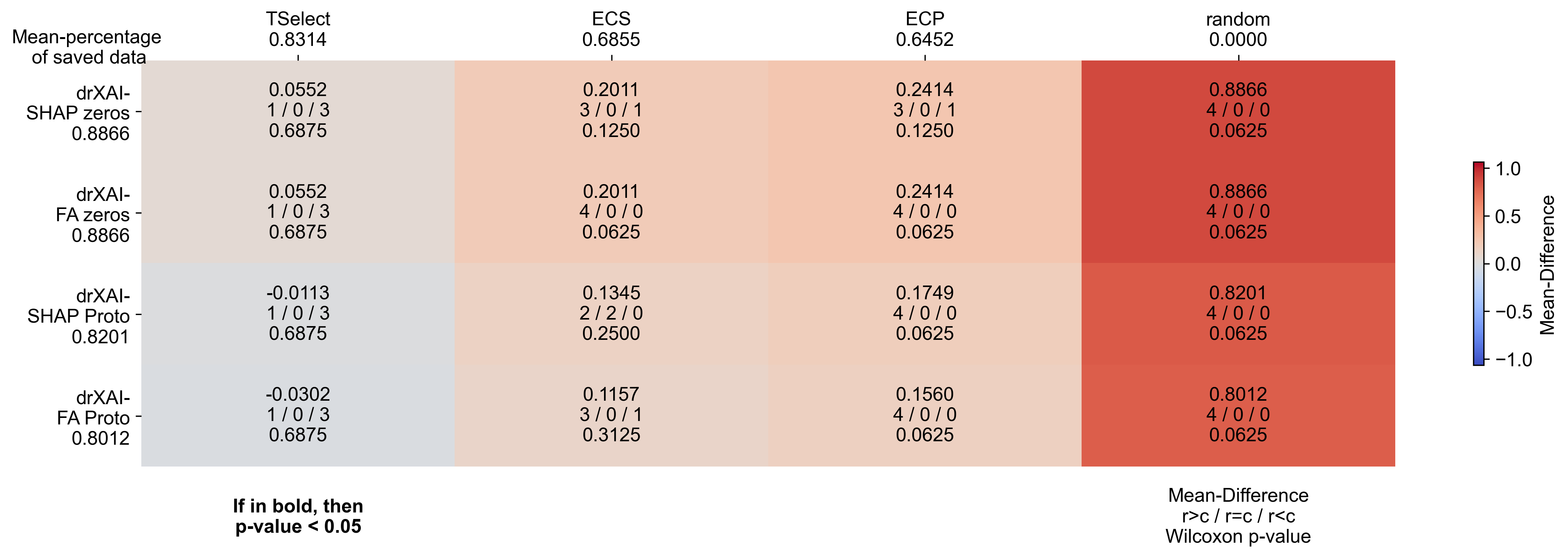}
    \caption{Mean percentage of data saved by each selection for the 4 MTSC datasets.}
    \label{fig:multi_savedData}
\end{figure}

For real-world datasets, we focus on those with a large number of channels and time points, where the benefits of feature selection are more pronounced. In order to test drXAI on large-scale MTSC datasets, we selected Face Detection \cite{olivetti2014meg} (144 channels, 62 time points), Arc Loss \cite{arc_loss} (96 channels, 1101 time points), Military Press and Rowing (mean-centered datasets, 50 channels and 161 time points each) \cite{ashishecml23}.  \\
We summarize the findings using \textit{Multi Comparison Matrices} (MCM) \cite{ismail2023approach} for mean accuracy and for percentage of data saved, respectively, in Figure \ref{fig:multi_acc} and \ref{fig:multi_savedData}. 

Figure \ref{fig:multi_acc} highlights that channel reduction is an important task, as 3 configurations of our method and 2 baselines have a higher average accuracy than using all features in the dataset. 
Our most accurate configuration, drXAI-SHAP-Proto, stands between the best 2 baselines accuracies, i.e., ECP and ECS. Using the faster FA explainer, our proposed background, Proto, is worse than the zero background, mainly due to poor performance when coupled with the MP dataset (see Appendix), although it is better than both ECP and ECS on 3 out of 12 experiments. TSelect has the lowest accuracy among the compared methods. \\
Analyzing the percentage of saved data (i.e., how many channels are discarded), Figure \ref{fig:multi_savedData} shows that \textbf{each configuration of our method considerably saves more data} than ECP and ECS, with at least $80\%$ reduction. In this regard, the zero-background achieves a better reduction than Proto. \\
Globally, \textbf{our method is the best trade-off between accuracy and data saved}, as different configurations achieve very good performance in both aspects, while ECP and ECS excel only in accuracy and TSelect in data reduction.

\subsection{Time Point Selection on UTSC Datasets}
\label{sec:utsc_exp}

Since UTSC datasets have more samples than the MTSC ones (see Appendix), we empirically set $n_{samp}$ to max 100 samples per class, rather than 50.  
We also decrease the number of non-improving epochs before early stopping to 10. \\
Lastly, since defining each time point as a feature in $F_{sel}$ would be extremely expensive for computing attribution, we instead group them in 20 consecutive, equal-length windows as done in \cite{serramazza2024improving}. This means that all time points within a window are either all selected or all discarded.

The baselines considered in this section are: the filter method Mutual Information (MI) \cite{scikit-learn}, the wrapper method Random Forest feature importance (RFI), and random selection. Similar to the channel selection procedure, random selection samples both the number and the specific time points from a uniform distribution.
Other possible baselines, such as recursive feature elimination, are not considered due to their high computational cost, especially for datasets with long time series.


\subsubsection{Synthetic Dataset.}

\begin{table} [t]
    \centering
     \resizebox{0.99\textwidth}{!}{
    \begin{tabular}{|l|c|c|c|c|c|c|c|}
     \hline
     
      & drXAI-FA-Proto & drXAI-SHAP-Proto & drXAI-FA-zeros & drXAI-SHAP-zeros  & RFI & MI & random \\
     \hline
         n. informative 10k  & 9,000  & \textbf{10,000} &  6000  & 3000 & 3252 & 1767 & 6,000 \\
    \hline
        n. uninformative 10k  & \textbf{0} & \textbf{0} & \textbf{0} & \textbf{0} &  \textbf{0} & \textbf{0} &2,000 \\
    \hline
      mean accuracy & .850 & \textbf{0.882} & .801 & .608 & .694 & .749 & 0.627 \\
    \hline
    \end{tabular}
    }
    \caption{Number of informative and uninformative time points for synthetic UTSC data and relative mean accuracy of compared feature selection methods (across the 3 runs of each of the 3 SOTA  classifiers).}
    \label{tab:synth_uni}
\end{table}

For the UTSC synthetic data, we used the data generator code included in \cite{nguyen2025tshap}, mirroring the MTSC case, but with $d=1$ channels.  
Two support waves, injected in random places within the first 10,000 \textit{informative} time points, define the same binary task as in the MTSC case. This informative area is followed by another 10,000 \textit{uninformative} points.
In this case, since using such a long series has a severe consequence on the running times of models, we used only 1,000 samples for training, and kept 5,000 for the test set. As in the former MTSC case, ideally, the first half of the features is selected, while the remaining second is discarded. 

Table \ref{tab:synth_uni} shows that each evaluated method only selects from relevant time points. Among our configurations, \textbf{drXAI-SHAP-Proto has a perfect selection}, selecting  all 10,000 important features; drXAI-FA-Proto selected 9,000. Among the remaining methods, only drXAI-FA-Zeros, selecting 6,000 features, achieves a good accuracy, since the baselines RFI and MI, as well as SHAP-Zero, select at most 3,252 features, retaining insufficient signal for accurate classification. \\
ConvTran cannot be trained using all features due to exceeding  the GPU VRAM available. Nevertheless, it can run on reduced data of each selection: using drXAI-SHAP-Proto, it achieves a 0.921 average accuracy (Appendix).

\subsubsection{Real-world Data.}

\begin{figure} [t]
    \centering
    \includegraphics[width=1\linewidth]{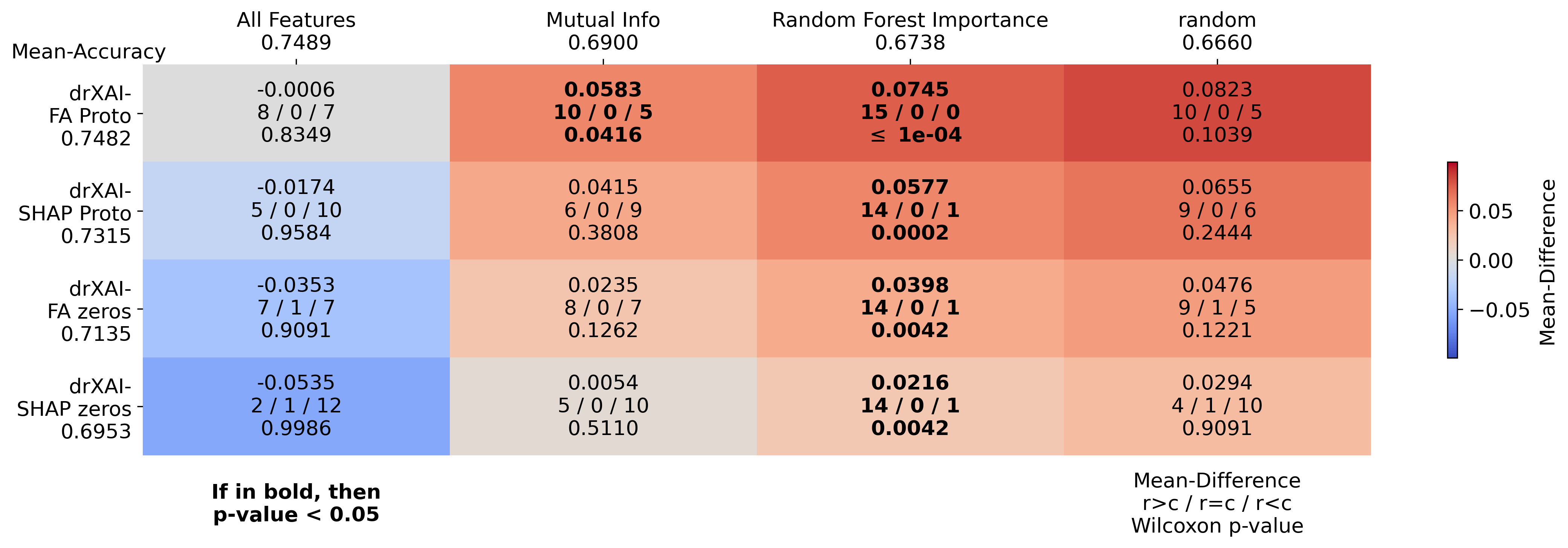}
    \caption{Mean accuracy of each selection (and All Features) for the 5 UTSC datasets used and the 3 classifiers, yielding 15 results in total.}
    \label{fig:uni_acc}
\end{figure}

\begin{figure} [t]
    \centering
    \includegraphics[width=1\linewidth]{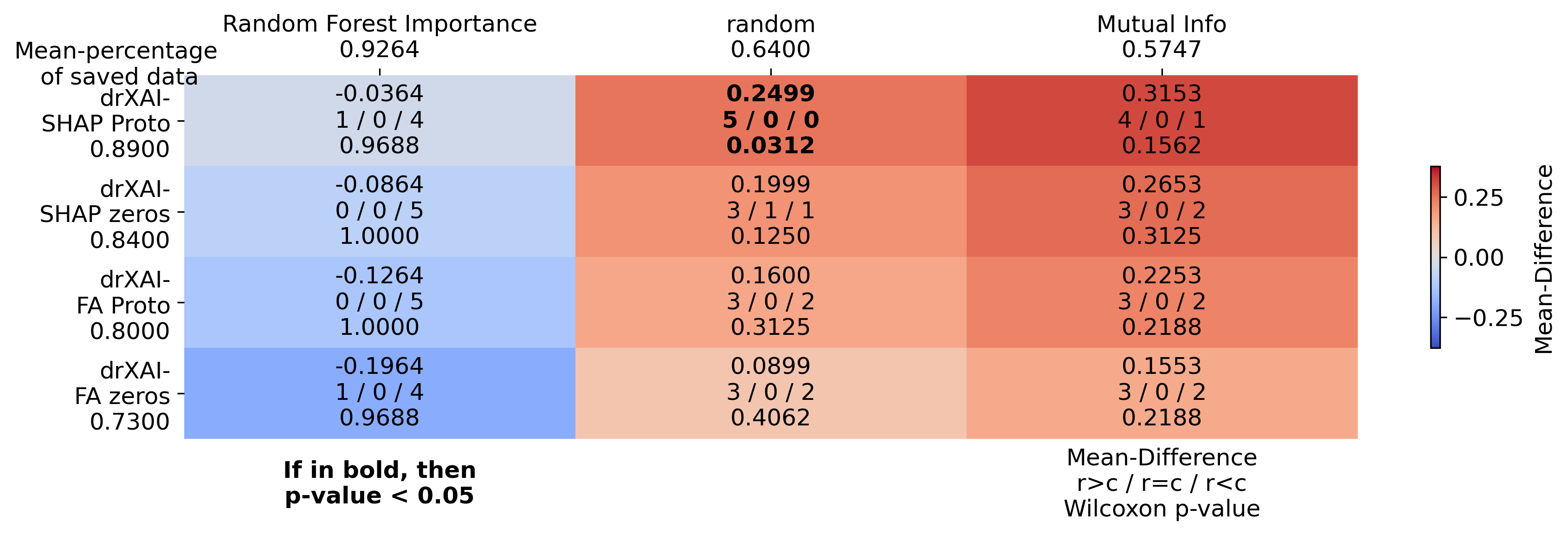}
    \caption{Mean percentage of data saved of each selection for the 5 UTSC datasets.}
    \label{fig:uni_savedData}
\end{figure}

For real-world datasets, we used large-scale datasets Cornell Whale Challenge, Mosquitos Sound and Whale Sounds from the MONSTER benchmark  \cite{dempster2025monstermonashscalabletime}, Right Whale Calls \cite{cox2006understanding} and Urban Sound \cite{salamon2014dataset}.  These datasets were chosen due to the number of time points (2.5-44k length) and samples (2.7-84k). \\
We note that using the Urban Sound dataset, which has 44k time points, ConvTran can be trained only using the reduced data after feature selection, as it otherwise exceeds the GPU memory limit. \\
Figure \ref{fig:uni_acc} and \ref{fig:uni_savedData} show, respectively, the MCM for accuracy across all datasets and classifiers and the MCM for percentages of data saved of each selection. In this case, \textbf{all configurations of our method achieve higher mean accuracy compared to the baselines}. The most accurate configuration is drXAI-FA-Proto, 
having accuracy comparable to the original datasets using all features. Configurations using the informative Proto background have a big margin over those using zeros. Comparing the baselines, MI outperforms RFI. 
Focusing on data saved, RFI has the largest average save, followed by our method, and lastly MI. \\
Among our configurations, the more expensive and accurate SHAP can save more data than the FA, and for both explainers, the Proto background saves more data than the zero background. We note that our configurations, excluding FA Zeros, save at least $80\%$ of the data. 
As with MTSC, drXAI is the only method achieving a good trade-off between accuracy and data reduction. RFI reduces data aggressively but at the cost of the worst accuracy, while MI shows no clear advantage in either aspect.

\subsection{Accuracy-Time Trade-off Analysis}
\begin{figure}[htb]
    \centering
    \begin{subfigure}[b]{0.225\textwidth}
        \includegraphics[width=\textwidth]{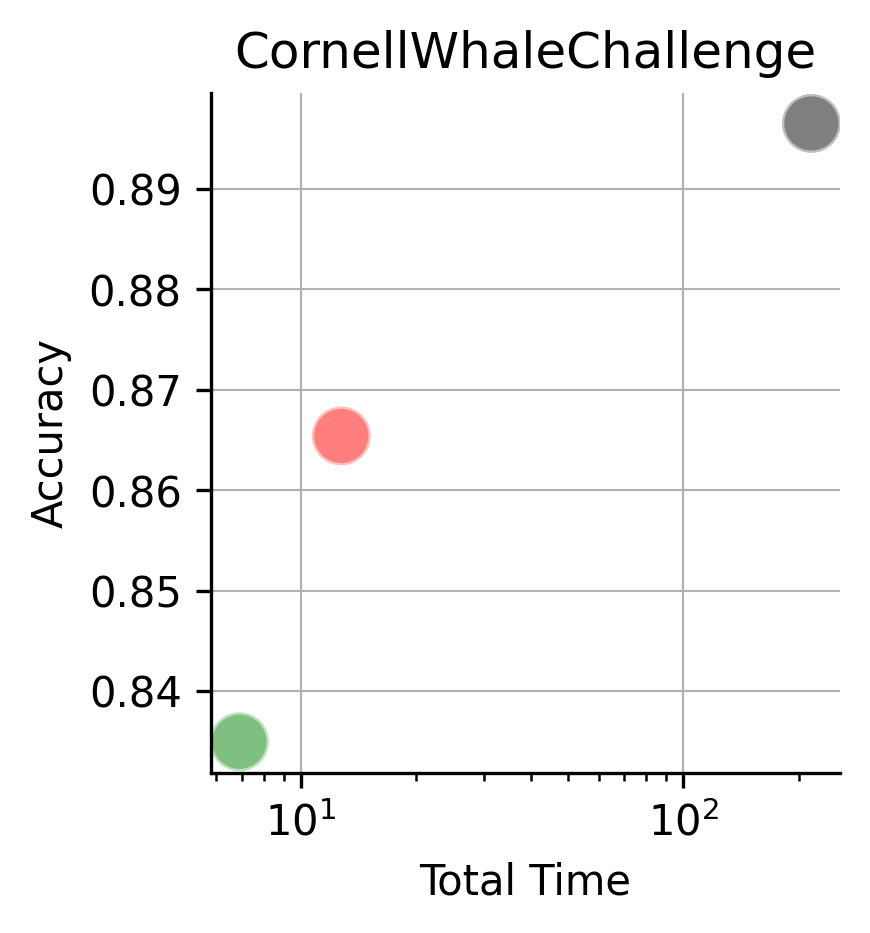}
    \end{subfigure}
    \begin{subfigure}[b]{0.225\textwidth}
        \includegraphics[width=\textwidth]{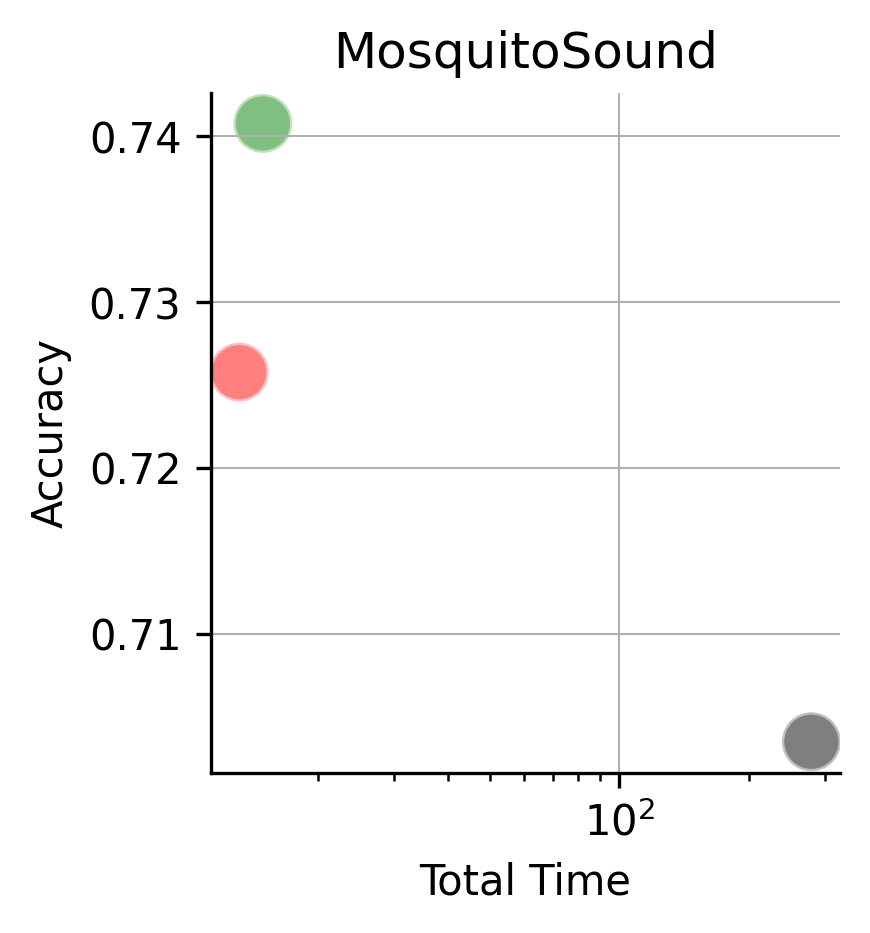}
    \end{subfigure}
    \begin{subfigure}[b]{0.225\textwidth}
        \includegraphics[width=\textwidth]{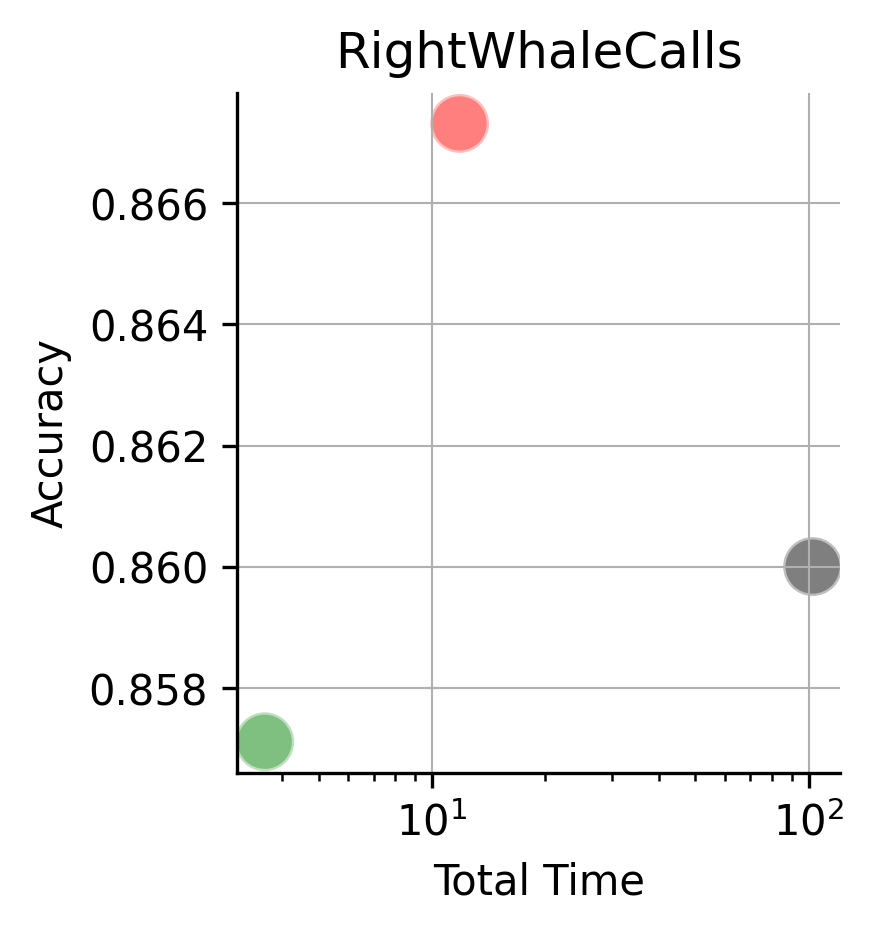}
    \end{subfigure}
    \begin{subfigure}[b]{0.3\textwidth}
        \includegraphics[width=\textwidth]{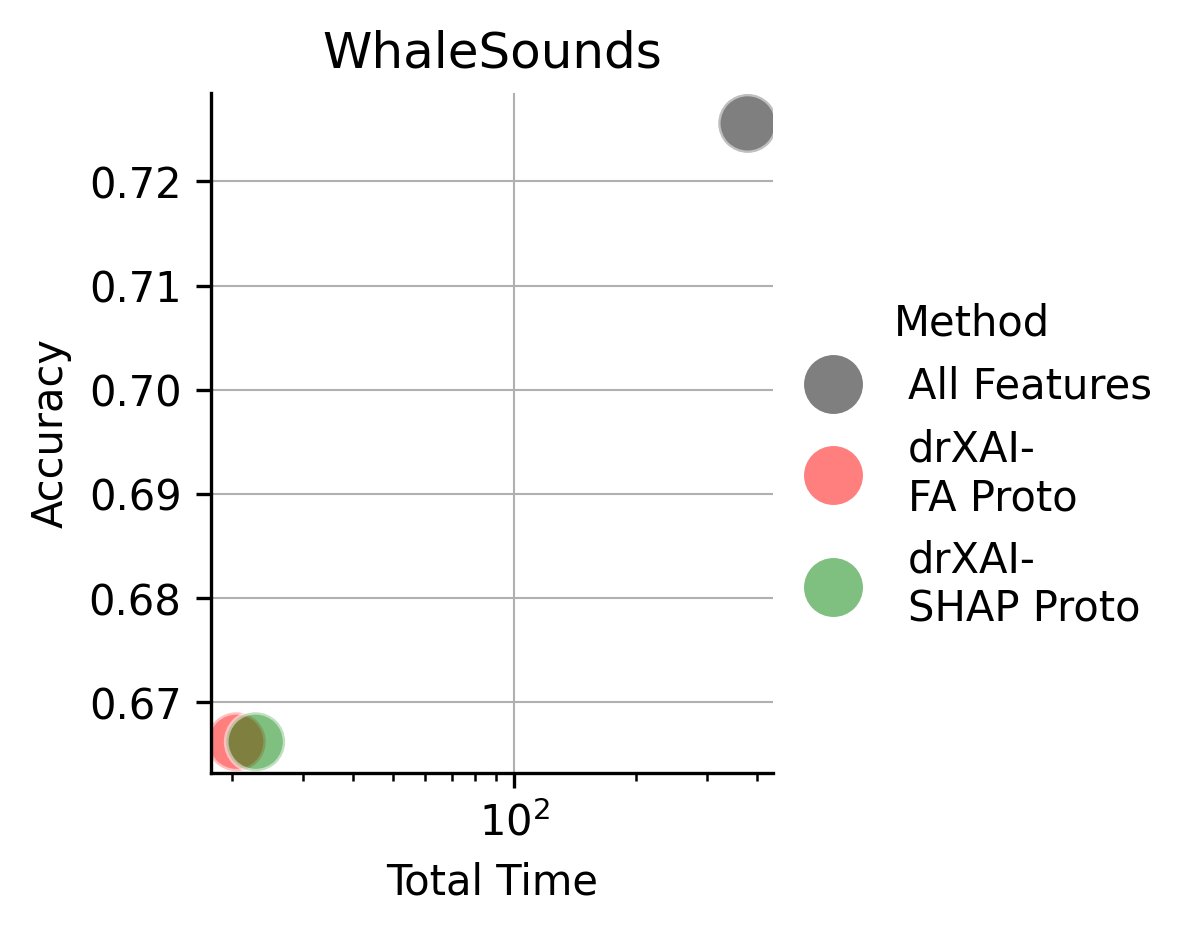}
    \end{subfigure}
    \caption{Accuracy vs Time (minutes in log scale) for ConvTran on UTSC datasets. Total time compares training on All Features vs drXAI pipeline (Hydra training + explanation + training on reduced data). On 2 of 6 datasets, All Features cannot run due to exceeding GPU memory.}
    \label{fig:accs_vs_DataSaved}
\end{figure}

An aspect worth analyzing is the total time required by drXAI compared to directly training the model using the original data. In our method, in addition to the cost of training the SOTA classifiers using the reduced data, the time to train Hydra and to compute the explanation (and thus the selection) must be considered. Figure \ref{fig:accs_vs_DataSaved} shows the accuracy vs total training time of 2 drXAI configurations vs  \textit{All Features} (training on non-reduced data) for ConvTran (the most expensive classifier), on 4 of the UTSC datasets, where the model can run using \textit{All Features}. The plots show that the accuracy is comparable, while the total time is reduced by one order of magnitude. The Appendix shows a detailed analysis of this cost.

Overall, based on our experiments, SHAP is preferable when aggressive data reduction is the priority, while FA is the best choice under time constraints, offering a faster yet effective selection.

\section{Conclusion}

This paper introduces drXAI, a novel methodology that repurposes XAI attribution methods to drive effective data reduction in TSC. By leveraging the GPU-accelerated Hydra classifier and fast explainers, we successfully bridge the gap between XAI interpretability and practical model scalability. Our framework leverages two complementary aggregation strategies (absolute and average), identifying the most salient features using an automated elbow-cut heuristic, avoiding the need for manual thresholds.

We validated drXAI across large-scale synthetic and real-world benchmarks using three SOTA classifiers: MultiRocket-Hydra, InceptionTime, and ConvTran. Our results demonstrate that drXAI consistently achieves over 80-90\% data reduction while maintaining, and in some cases exceeding, classification accuracy on the full dataset. Specifically, drXAI-SHAP-Proto emerged as the most accurate configuration for channel selection in MTSC, while drXAI-FA-Proto led in performance for time-point selection in UTSC. The proposed Proto background marginally outperforms the zero background, at no additional cost; SHAP  achieves superior data reduction compared to Feature Ablation but at the expense of longer computation time.

Crucially, we demonstrate that data reduction enabled by the lightweight Hydra model allows resource-intensive architectures like ConvTran to scale to massive datasets that were previously inaccessible due to GPU memory constraints. While limitations exist regarding the computational overhead of SHAP on extremely long sequences, by grouping features drXAI shows that XAI can be a robust, flexible tool for high-performance feature selection. Future work will extend this framework to regression tasks and explore the simultaneous reduction of both channels and time points.

\section*{Acknowledgments}
This publication has emanated from research supported in part by a grant from Science Foundation Ireland under Grant number 18/CRT/6183. For the purpose of Open Access, the author has applied a CC BY public copyright licence to any Author Accepted Manuscript version arising from this submission.

\bibliographystyle{abbrv}  
\bibliography{bibliography} 

\end{document}


\section*{Appendix drXAI}

\subsection*{Dataset Characteristics}
\begin{table}[]
    \centering
    \begin{tabular}{|c|c|c|c|c|c|}
         \hline
         dataset name & n. samples train & n. samples test & n. channels & n. time points & n. classes \\
         \hline
         synthetic MTSC & 5000 & 5000 & 40 & 1000 & 2 \\ 
        Arc Loss 
        & 2581 & 645 & 96 & 1101 & 2 \\ 
        FaceDetection & 5890 & 3524 & 144 & 62 & 2 \\ 
        MP & 1426 & 595 & 50 & 161 & 4 \\ 
        Rowing & 1838 & 790 & 50 & 161 & 5 \\ 
       
        \hline
         synthetic UTSC & 1000 & 5000 & 1 & 20000 & 2 \\ 
        CornellWhaleChallenge & 24000 & 6000 & 1 & 4000 & 2 \\ 
        MosquitoSound & 55913 & 223653 & 1 & 3750 & 6 \\ 
        RightWhaleCalls & 10934 & 1962 & 1 & 4000 & 2 \\ 
       
        UrbanSound & 2717 & 2718 & 1 & 44100 & 10 \\ 
        WhaleSounds & 84131 & 21032 & 1 & 2500 & 8 \\
            \hline
    \end{tabular}
    \caption{Characteristics of all datasets used in the paper.}
\end{table}

\newpage
\subsection*{MTSC Accuracy}

\begin{table}[]
    \centering
    \begin{tabular}{|c|c|c|c|c|c|c|c|c|c|c|}

        \hline
        dataset & classifier & \makecell{All \\ Features} & \makecell{FA \\ Proto} & \makecell{SHAP \\ Proto} & \makecell{FA  \\ zeros} & \makecell{SHAP \\ zeros} & ECP & ECS & TSelect & random \\
        \hline
        Arc Loss & ConvT. &  .748 &  .749 &  \textbf{.754} &  .740 &  .744 &  .745 &  .753 &  .671 & .725 \\ 
        Arc Loss & IncepT. &  .726 &  .716 &  .719 &  .729 &  .709 &  .724 &  \textbf{.736} &  .654 & .678 \\ 
        Arc Loss & MRH &  \textbf{.740} &  .734 &  .726 &  .726 &  .730 &  .729 &  .733 &  .644 & .674 \\ 
        \hline
        Face D.& ConvT. &  .560 &  .570 &  .593 &  .596 &  .559 &  .603 &  \textbf{.621} &  .564 & .566 \\ 
        Face D.& IncepT. &  \textbf{.677} &  .589 &  .598 &  .610 &  .599 &  .634 &  .639 &  .659 & .663 \\ 
        Face D.& MRH &  \textbf{.602} &  .553 &  .561 &  .563 &  .551 &  .594 &  \textbf{.602} &  .599 & .592\\ 
        \hline
        Military Press & ConvT. &  .538 &  .599 &  .683 &  \textbf{.733} &  .707 &  .638 &  .648 &  .243 & .509\\ 
        Military Press & IncepT. &  .661 &  .527 &  .641 &  .604 &  .575 &  \textbf{.665} &  .555 &  .261 & .545\\ 
        Military Press & MRH &  .7 &  .768 &  .82 &  .807 &  .787 &  \textbf{.833} &  .78 &  .248 & .648\\ 
        \hline
         Rowing & ConvT. &  .671 &  .774 &  \textbf{.800} &  .758 &  .71 &  .77 &  .752 &  .553 & .547 \\ 
        Rowing & IncepT. &  .605 &  .672 &  .669 &  .675 &  .6 &  .689 &  \textbf{.729} &  .567 & .434\\ 
        Rowing & MRH &  \textbf{.755} &  .729 &  .744 &  .725 &  .719 &  .737 &  .705 &  .568 & .688\\ 
        \hline
        Synthetic MTSC & ConvT. &  .828 &  \textbf{.858} &  .693 &  .597 &  .568 &  .714 &  .597 &  .589 & .53 \\ 
        Synthetic MTSC & IncepT. &  .518 &  .527 &  \textbf{.653} &  .589 &  .551 &  .511 &  .572 &  .524 & .529 \\ 
        Synthetic MTSC & MRH &  \textbf{.830} &  .820 &  .628 &  .571 &  .547 &  .677 &  .566 &  .552 & .53\\
        \hline
    \end{tabular}
    \caption{MTSC accuracies. Face D. stands for Face detection, ConvT. for ConvTran and IncepT. for InceptionTime.}
\end{table}

\subsection*{MTSC Data Selection Details}

\begin{table}[]
    \centering
    \begin{tabular}{|c|c|c|c|c|c|c|c|c|c|c|c|c|c|c|c|c|}
        \hline
        dataset &  \multicolumn{2}{c|}{FA Proto} & \multicolumn{2}{c|}{SHAP Proto}  & \multicolumn{2}{c|}{FA zeros}  & \multicolumn{2}{c|}{SHAP zeros}   & \multicolumn{2}{c|}{ECP} &  \multicolumn{2}{c|}{ ECS} & \multicolumn{2}{c|}{TSelect} & \multicolumn{2}{c|}{random}  \\
        \hline
          & n & \% & n & \% & n & \% & n & \% & n & \% & n & \% & n & \%  & n & \% 15 \\
        \hline
        Arc Loss & 14 & 0.146 & 12 & 0.125 & 12 & 0.125 & 8 & 0.083 & 38 & 0.396 & 59 & 0.615 & 1 & 0.01 & 15 & 0.16 \\ 
        FaceDetection & 10 & 0.069 & 5 & 0.035 & 7 & 0.049 & 13 & 0.09 & 12 & 0.083 & 12 & 0.083 & 87 & 0.604 & 117 & 0.81\\ 
        Rowing & 11 & 0.22 & 14 & 0.28 & 7 & 0.14 & 9 & 0.18 & 22 & 0.44 & 14 & 0.28 & 1 & 0.02 & 32 & 0.64 \\ 
        MP & 18 & 0.36 & 14 & 0.28 & 7 & 0.14 & 5 & 0.1 & 25 & 0.5 & 14 & 0.28 & 2 & 0.04 & 4 & 0.08 \\
    \hline
    \end{tabular}
    \caption{Selection length for real-world MTSC datasets. For each selection of our method and baselines, number of selected channels (n) and relative percentage of the total. }
\end{table}

\newpage

\subsection*{UTSC Accuracy}
\begin{table}[]
    \centering
    \begin{tabular}{|c|c|c|c|c|c|c|c|c|c|}
        \hline 
         dataset & classifier & \makecell{All \\ Features} & \makecell{FA \\ Proto} & \makecell{SHAP \\ Proto} & \makecell{FA  \\ zeros} & \makecell{SHAP \\ zeros} & RFI & MI & random \\
         \hline
        CornellWhaleChallenge & ConvT. & \textbf{.897} & .865 & .835 & .872 & .805 & .778 & .892 & .827 \\ 
        CornellWhaleChallenge & IncepT. & .863 & .867 & .843 & \textbf{.875} & .813 & .783 & .867 & .835 \\ 
        CornellWhaleChallenge & MRH & .\textbf{838} & .807 & .756 & .825 & .759 & .727 & .832 & .757 \\ 
        \hline
        MosquitoSound & ConvT. & .703 & .726 & \textbf{.741} & .714 & .726 & .684 & .577 & 0.66\\ 
        MosquitoSound & IncepT. & \textbf{.930} & .819 & .843 & .783 & .843 & .756 & .658 & .892 \\ 
        MosquitoSound & MRH & \textbf{.885} & .771 & .794 & .727 & .786 & .691 & .537 & .838 \\ 
        \hline
        RightWhaleCalls & ConvT. & .860 & .867 & .857 & \textbf{.878} & .792 & .739 & .867 & .831\\ 
        RightWhaleCalls & IncepT. & .846 & \textbf{.871} & .854 & .850 & .784 & .740 & .850 & .811 \\ 
        RightWhaleCalls & MRH & .810 & .814 & .781 & \textbf{.819} & .711 & .688 & .799 & .753\\ 
        \hline
        WhaleSounds & ConvT. & \textbf{.726} & .666 & .666 & .666 & .666 & .644 & .672 & .694 \\ 
        WhaleSounds & IncepT. & .691 & .703 & .703 & .703 & .704 & .681 & \textbf{.713} & .727 \\ 
        WhaleSounds & MRH & .637 & .638 & .639 & .639 & .637 & .607 & \textbf{.643} & .634\\ 
        \hline
        UrbanSound & ConvT. & NA & \textbf{.499} & .437 & NA & NA & .459 & .444 & NA \\ 
        UrbanSound & IncepT. & \textbf{.729} & .627 & .588 & .645 & .674 & .571 & .484 & NA \\ 
        UrbanSound & MRH & \textbf{.817} & .683 & .636 & .707 & .731 & .559 & .515 & .732\\ 
        \hline
        synthetic UTSC & ConvT. & NA & .838 & \textbf{.921} & .710 & .608 & .598 & .605 & .647 \\ 
        synthetic UTSC & IncepT. & .508 & .813 & .647 & \textbf{.828} & .589 & .673 & .792 & .571 \\ 
        synthetic UTSC & MRH & \textbf{.961} & .898 & .957 & .865 & .626 & .810 & .851 & .664\\
        \hline
    \end{tabular}
    \caption{UTSC accuracies. Experiments where the method exceeded the available memory are denote with NA. ConvT. stands for ConvTran, IncepT. stands for InceptionTime.}
\end{table}

\newpage

\subsection*{UTSC Data Selection Details}

    \begin{table}[]
        \centering
    \begin{tabular}{|c|c|c|c|c|c|c|c|c|c|c|c|c|c|c|c|} 
            \hline
        dataset &  \multicolumn{2}{c|}{FA Proto} & \multicolumn{2}{c|}{SHAP Proto}  & \multicolumn{2}{c|}{FA zeros}  & \multicolumn{2}{c|}{SHAP zeros}   & \multicolumn{2}{c|}{RFI} &  \multicolumn{2}{c|}{ MI} &  \multicolumn{2}{c|}{ random } \\
        \hline
        & n  & \% & n  & \% & n  & \% & n  & \% & n  & \% & n  & \% & n  & \% \\ 
        \hline
        CornellWhaleChallenge & 1199 & 0.30 & 399 & 0.10 & 1200 & 0.30 & 800 & 0.20 & 341 & 0.09 & 3761 & 0.94 & 600 & 0.15 \\ 
        MosquitoSound & 376 & 0.10 & 564 & 0.15 & 188 & 0.05 & 564 & 0.15 & 208 & 0.06 & 104 & 0.03 & 2065 & 0.55 \\ 
        RightWhaleCalls & 1600 & 0.40 & 600 & 0.15 & 3000 & 0.75 & 599 & 0.15 & 355 & 0.09 & 3636 & 0.91 & 100 & 0.25 \\ 
        WhaleSounds & 250 & 0.10 & 250 & 0.10 & 250 & 0.10 & 250 & 0.10 & 203 & 0.08 & 408 & 0.16 & 1625 & 0.65 \\ 
        UrbanSound & 4410 & 0.10 & 2205 & 0.05 & 6615 & 0.15 & 8820 & 0.20 & 2526 & 0.06 & 3804 & 0.09 & 8820 & 0.2 \\

        \hline
        \end{tabular}
        \caption{Selection length for real-world UTSC datasets. For each selection of our method and baselines, number of selected channels (n) and relative percentage of the total}
    \end{table}

\newpage

\subsection*{Time analysis on UTSC data}

\begin{table}[]
    \centering
    \begin{tabular}{|c|c|c|c|c|c|c|c|c|c|c|c|c|c|c|}
        \hline
        dataset & classifier  & AF Time & \multicolumn{4}{c|}{drXAI-FA Proto} &  \multicolumn{3}{c|}{drXAI-SHAP Proto} \\
        \hline
          & & & hydra time & exp time & train time & TOTAL & exp time & train time & TOTAL \\ 
        \hline
        RWC & CON & 102.3 & 0.3 & 0.04 & 11.5 & 11.84 & 1.04 & 2.3 & 3.64 \\ 
        RWC & INC & 30.87 & 0.3 & 0.04 & 11.6 & 11.94 & 1.04 & 4.4 & 5.74 \\ 
        RWC & MRH & 5.6 & 0.3 & 0.04 & 2.4 & 2.74 & 1.04 & 1.3 & 2.64 \\ 
        \hline
        CWC & CON & 216.06 & 0.6 & 0.04 & 12.1 & 12.74 & 1.02 & 5.3 & 6.92 \\ 
        CWC & INC & 76.43 & 0.6 & 0.04 & 25.1 & 25.74 & 1.02 & 6.2 & 7.82 \\ 
        CWC & MRH & 15.39 & 0.6 & 0.04 & 7.3 & 7.94 & 1.02 & 5.1 & 6.72 \\ 
        \hline
        MS & CON & 278.56 & 3.3 & 0.13 & 9.7 & 13.13 & 3 & 8.6 & 14.9 \\ 
        MS & INC & 165.38 & 3.3 & 0.13 & 9.8 & 13.23 & 3 & 17.6 & 23.9 \\ 
        MS & MRH & 168.91 & 3.3 & 0.13 & 76.4 & 79.83 & 3 & 81 & 87.3 \\ 
        \hline
        WS & CON & 378.89 & 1.2 & 0.1 & 19.2 & 20.5 & 2.5 & 19.2 & 22.9 \\ 
        WS & INC & 201.49 & 1.2 & 0.1 & 11.4 & 12.7 & 2.5 & 11.2 & 14.9 \\ 
        WS & MRH & 179.71 & 1.2 & 0.1 & 140.8 & 142.1 & 2.5 & 145.1 & 148.8 \\ 
        \hline
        US & CON & NA & 2.5 & 6.51 & 48.6 & 57.61 & 150.64 & 11.3 & 164.44 \\ 
        US & INC & 195.57 & 2.5 & 6.51 & 15.4 & 24.41 & 150.64 & 10.4 & 163.54 \\ 
        US & MRH & 31.67 & 2.5 & 6.51 & 2.7 & 11.71 & 150.64 & 1.3 & 154.44 \\ 
        \hline
        SY & CON & NA & 1.1 & 0.65 & 93.1 & 94.85 & 14.49 & 88.4 & 103.99 \\ 
        SY & INC & 12.87 & 1.1 & 0.65 & 5.6 & 7.35 & 14.49 & 5.5 & 21.09 \\ 
        SY & MRH & 15.1 & 1.1 & 0.65 & 6.4 & 8.15 & 14.49 & 7.2 & 22.79 \\
        \hline
    \end{tabular}
    \caption{Detailed time comparison (in minutes) among all features (AF TIME), drXAI-FA Proto and drXAI-SHAP Proto, across the 3 SOTA classifiers ConvTran (CON), InceptionTime (INC) and MultiRocket-Hydra(MRH) using the UTSC dataset RightWhaleCalls (RWC), CornellWhaleChallenge (CWC), MosquitoSound (MS), WhaleSounds (WS), UrbanSound (US), and synthetic UTSC (SY). To have a lighter table, we omit the Hydra time For drXAI-SHAP Proto , as it is identical to the one for drXAI-FA Proto.}
\end{table}

\begin{table}[]
    \centering
    \begin{tabular}{|c|c|c|c|}
        \hline
        dataset & classifier & drXAI-FA Proto time/AF Time & drXAI-SHAP Proto time/AF Time \\
        \hline
        RightWhaleCalls & ConvT. & 0.116 & 0.036\\ 
        RightWhaleCalls & IncepT. & 0.387 & 0.186\\ 
        RightWhaleCalls & MRH & 0.489 & 0.471\\ 
        \hline
        CornellWhaleChallenge & ConvT. & 0.059 & 0.032\\ 
        CornellWhaleChallenge & IncepT. & 0.337 & 0.102\\ 
        CornellWhaleChallenge & MRH & 0.516 & 0.437\\ 
        \hline
        MosquitoSound & ConvT. & 0.047 & 0.053\\ 
        MosquitoSound & IncepT. & 0.080 & 0.145\\ 
        MosquitoSound & MRH & 0.473 & 0.517\\ 
        \hline
        WhaleSounds & ConvT. & 0.054 & 0.060\\ 
        WhaleSounds & IncepT. & 0.063 & 0.074\\ 
        WhaleSounds & MRH & 0.791 & 0.828\\ 
        \hline       
        UrbanSound & ConvT. & NA & NA \\
        UrbanSound & IncepT. & 0.125 & 0.836\\ 
        UrbanSound & MRH & 0.370 & 4.877\\ 
        \hline
        synthetic UTSC & ConvT. &  NA & NA \\ 
        synthetic UTSC & IncepT. & 0.571 & 1.639\\ 
        synthetic UTSC & MRH & 0.540 & 1.509 \\
        \hline
    \end{tabular}
    \caption{Proportion between the total time of drXAI-FA Proto and All Features time (AF Time) as well as the proportion between the total time of drXAI-FA SHAP and All Features time for each SOTA classifier and for each UTSC dataset. We denote with NA where the classifier can't run using all features due to GPU exceeding the available memory.}
\end{table}